\documentclass[11pt]{article}
\usepackage[margin=1in]{geometry}
\usepackage{amsmath,amssymb,amsthm,mathtools,bm,dsfont}
\usepackage{algorithm,algorithmicx,algpseudocode}
\usepackage{booktabs}
\usepackage{authblk}
\usepackage{multirow}
\usepackage{microtype}
\usepackage{upgreek}
\usepackage{lmodern}
\usepackage{xcolor}
\usepackage{etoc}
\usepackage{graphicx}
\usepackage{float}
\usepackage[font=footnotesize, margin=1cm]{caption}

\usepackage{natbib}
\bibpunct[, ]{(}{)}{,}{a}{}{,}%

\usepackage[hidelinks]{hyperref}
\hypersetup{
	colorlinks=false
}

\newcommand{\ind}{\mathds{1}}
\newcommand{\Ans}{\mathcal{A}^{\star}}
\newcommand{\R}{\mathbb{R}}
\newcommand{\X}{\mathcal{X}}

\newcommand{\E}{\mathbb{E}}
\newcommand{\Prob}{\mathbb{P}}

\DeclareMathOperator*{\argmax}{argmax}
\DeclareMathOperator*{\argmin}{argmin}
\newcommand\name[1]{\textup{\texttt{#1}}}

\newtheorem{proposition}{Proposition}

\newtheorem{remark}{Remark}
\newtheorem{example}{Example}

\title{Ranking-and-Selection with Multiple Correct Answers and Non-Answerable Estimates}
\author[1]{Qiaoqiao Wang}
\author[1]{Wei You}
\affil[1]{Dept.~of Industrial Engineering and Decision Analytics, The Hong Kong University of Science and Technology, Hong Kong, China}

\begin{document}

\maketitle

\begin{abstract}
  We study fixed-precision ranking-and-selection in structured settings where the answer may be non-unique and where noisy estimates may temporarily admit no valid answer at all.
  This phenomenon arises naturally in problems such as multi-fidelity ranking-and-selection and identifying a Condorcet winner from pairwise comparisons.
  To address this, we propose a unified framework based on answer-wise acceptance sets, restricted generalized likelihood ratio stopping, and an answer--pitfall decomposition that yields a max-max-min characteristic value and a common sampling principle.
  We introduce ENDS, a general procedure that combines estimation, nomination, pitfall detection, and cost-aware information-directed selection.
  We instantiate ENDS for various problems by deriving explicit formulas.
  Extensive numerical experiments show that this unified recipe performs well across a broad range of pure-exploration problems and offers a practical framework and proof-of-concept algorithmic recipe.
\end{abstract}

\section{Introduction}

Many ranking-and-selection (R\&S) problems arise in settings where information is noisy, structured, and expensive.
In multi-fidelity experimentation, one can query cheap but biased proxies or expensive high-fidelity measurements; in dueling bandits, feedback arrives only through pairwise comparisons rather than direct rewards.
These models are increasingly natural in engineering design, simulation optimization, preference learning for LLMs, and human-in-the-loop evaluation, where absolute scores are often unavailable or prohibitively costly and decisions must be made with a prescribed level of confidence.

What makes these settings especially challenging is that the usual single-winner template is no longer sufficient.
First, the answer map may be set-valued: in good-alternative or subset-selection problems, several answers can be simultaneously correct.
Second, even when the true instance is answerable, a noisy estimate may temporarily fall outside the answerable set.
In multi-fidelity ranking-and-selection, empirical means may violate the bias constraints; in dueling bandits, estimated preferences may induce ties or cycles so that no Condorcet winner exists.
In such situations, the empirical-best answer can be undefined exactly when the algorithm still needs a principled way to sample, stop, and decide.

The fixed-confidence pure-exploration literature has developed sharp lower bounds and asymptotically efficient procedures for best-arm identification, and has gradually expanded to structured settings such as multiple correct answers, multi-fidelity observations, and pairwise preference feedback \citep{garivier2016optimal,degenne2019pure,poiani2024optimal,chen2020combinatorial}.
In parallel, the ranking-and-selection literature has long emphasized the close connection between fixed-precision procedures and sequential hypothesis testing, and has developed important variants around good selection, subset selection, multi-fidelity simulation, and pairwise-comparison settings \citep{wang2024bonferroni,HongFanLuo2021Review,eckman2018guarantees,XuEtAl2016MO2TOS,ren2020sample}.
Related questions also arise in economics \citep{ramdas2020admissible,kasy2021adaptive}, where sequential experimentation, information acquisition, and stopping are studied under objectives that differ from bandit sample complexity but face many of the same conceptual tensions between exploration, confidence, and irreversible choice.

Despite this progress, the existing treatments remain largely problem-specific.
Standard formulations typically assume either a unique correct answer or an always-answerable estimate, and therefore do not directly address the joint presence of multiple correct answers and non-answerable estimates.
This gap matters algorithmically: it affects how one defines the stopping statistic, how one nominates a provisional answer, and how one allocates sampling effort when no empirical winner is currently well defined.

This paper takes a step toward that unification.
Building on the pitfall-adapted nomination perspective of \citet{qin2025dual}, we lift the framework from unique-answer settings to answer-wise acceptance sets and answer-wise pitfalls, so that the learner reasons over all candidate answers rather than only around a single true answer.
We view the paper as a proof of concept that a single structural recipe can transfer across broad models and problems.
The main contributions and organization of the paper are as follows.
\begin{itemize}
  \item In Section~\ref{sec:formulation}, we propose a unified framework for \textit{cost-aware} fixed-precision R\&S with a \textit{set-valued answer map} and an \textit{answerable region}.
  \item In Section~\ref{sec:algorithm}, we extend \citet{qin2025dual} with an answer-wise layer: rejection regions are decomposed by candidate answers and answer-wise pitfalls, yielding a restricted-GLR stopping rule and a max-max-min characteristic value that motivates the performance limit and the corresponding sampling principle.
  The proposed ENDS procedure combines estimation, answer-wise nomination, pitfall detection, and cost-aware information-directed selection in a portable way across problems.
  \item In Section~\ref{sec:algorithm}, we derive explicit instantiations for representative problems, including good-alternative selection, multi-fidelity ranking-and-selection, and Condorcet-winner dueling bandits.
  \item Through broad numerical experiments in Section~\ref{sec:numerical}, we show that the unified recipe is practically effective across qualitatively different exploration problems, supporting the case for a common algorithmic language beyond any one model class.
\end{itemize}

\section{Problem formulation}\label{sec:formulation}

Let $\Theta$ be the ambient space of valid parameters.
We allow $\Theta$ to be structured; see Example~\ref{ex:dueling}.
Let $\mathcal U$ be a finite set of measurement types.
At each round $n$, the learner chooses $U_n \in \mathcal U$ and observes $Y_n \sim P_{\bm\theta,U_n}$.
Each measurement type $u \in \mathcal U$ incurs measurement cost $C_u(\bm\theta) > 0$.
For every $u \in \mathcal U$, let $N_{n,u} \triangleq \sum_{\ell=1}^n \ind\{U_\ell = u\}$ and $p_{n,u} \triangleq N_{n,u}  / n$.
We write $\bm p_n = (p_{n,u})_{u \in \mathcal U}$ for the measurement allocation, and
\[
  \bar C_{\bm\theta}(\bm p) \triangleq \sum_{u \in \mathcal U} p_u C_u(\bm\theta)
\]
for the average cost per sample under allocation $\bm p$.
The cost spent after $n$ rounds is
\[
  B_n = \sum_{u \in \mathcal U} N_{n,u} C_u(\bm\theta) = n \bar C_{\bm\theta}(\bm p_n).
\]

The goal is to identify a correct answer from the set of possible answers $\mathcal A$, by allocating samples to the set of measurement types $\mathcal U$ and observing noisy measurements.
We allow the answer map to be set-valued.
For every parameter $\bm\theta \in \Theta$, we let $\Ans(\bm\theta) \subseteq \mathcal A$ denote the set of correct answers at $\bm\theta$.
In multiple-correct-answer problems, one can have $|\Ans(\bm\theta)| > 1$.

We say that an instance $\bm\theta$ is \textit{answerable} if $\Ans(\bm\theta) \neq \varnothing$.
Define the \textit{answerable set} as
\[
  \Theta^\star \triangleq \{ \bm\theta \in \Theta : \Ans(\bm\theta) \neq \varnothing \}.
\]
We require that the ground truth $\bm\theta \in \mathrm{interior}(\Theta^*)$, which excludes boundary instances at which arbitrarily small perturbations may change the set of correct answers, making exact identification statistically ill-posed.
To accommodate genuinely non-answerable instances, we may augment the answer set to include a null answer $a= \varnothing$, representing the conclusion that no substantive answer exists.
During learning the empirical estimate may leave the answerable set $\Theta^\star$ even if the true parameter does not, in which case an empirical answer need not exist.
We refer to such estimates as a \textit{non-answerable estimate}.
For each answer $a \in \mathcal A$, define its acceptance set by
\[
  \Theta^{\star}_a \triangleq \{ \bm\vartheta \in \Theta^\star : a \in \Ans(\bm\vartheta) \}.
\]
Let $\mathrm{Alt}_a \triangleq \Theta^\star \setminus \Theta^{\star}_a$ denote all answerable instances for which answer $a$ is not correct.

In this paper, we focus on the fixed-precision setting.
The goal is to design a stopping rule $\tau_\delta$ and a decision rule $\hat a_{\tau_\delta}$ such that
\[
  \Prob_{\bm\theta}(\hat a_{\tau_\delta} \in \Ans(\bm\theta)) \ge 1 - \delta
  \quad\text{for every }\bm\theta \in \Theta^\star,
\]
while designing a sampling rule that minimizes the expected total cost $\E_{\bm\theta}[B_{\tau_\delta}]$.

\subsection{Examples}
\begin{example}[Selection of a good alternative]\label{ex:good-alternative}
  Fix a tolerance $\varepsilon > 0$.
  We consider selection of a good alternative from a set of $K$ alternatives \citep{eckman2018guarantees}.
  This is also referred to as $\varepsilon$-best-arm identification \citep{jourdan2023varepsilon} or $(\varepsilon,\delta)$-PAC learning.
  Each measurement type is an alternative, and $\mathcal U=[K]$.
  Sampling alternative $i$ produces a noisy observation with mean $\theta_i$, and incurs cost $C_i(\bm\theta)$, where $\bm\theta=(\theta_i)_{i\in[K]}$.
  The ambient parameter is $\Theta$, the set of all possible mean vectors.
  For example, if the measurements are Gaussian, then the ambient parameter space is $\Theta = \R^{K}$.
  The set of possible answers is $\mathcal A=[K]$, and the set of correct answers is
  \[
  \Ans(\bm\theta)
  \triangleq
  \bigl\{
  i\in[K]: \theta_i \ge \max_{j\in[K]} \theta_j - \varepsilon
  \bigr\}.
  \]
  Thus every instance is answerable, so $\Theta^\star=\Theta$.
  This problem therefore exhibits multiple correct answers but no non-answerable estimates.
  The acceptance set for answer $i$ is
  \[
  \Theta_i^\star
  \triangleq
  \left\{
  \bm\vartheta \in \Theta : \vartheta_i \ge \vartheta_j - \varepsilon,\ \forall j\neq i
  \right\}.
  \]
\end{example}

\begin{example}[Multi-fidelity ranking-and-selection]\label{ex:MF-BAI}
  In multi-fidelity ranking-and-selection/BAI, each alternative $i\in[K]$ can be sampled at several fidelity levels $m\in[M]$ with fidelity-dependent costs $C_m$, with higher fidelities being more expensive and less biased \citep{poiani2024optimal}.
  The measurement set is then $\mathcal U = [K] \times [M]$, with measurement $u = (a,m)$ representing that the learner samples alternative $a$ at fidelity level $m$.
  The ambient parameter is $\bm\theta = (\theta_{a,m})_{a \in [K], m \in [M]} \in \Theta$, where $\Theta$ denotes the set of all possible mean tables.
  The multi-fidelity model class $\Theta_{\mathrm{MF}}$ consists of all mean tables that satisfy the multi-fidelity constraints $|\theta_{i,m} - \theta_{i,M}| \le \xi_m$, i.e.,
  \[
  \Theta_{\mathrm{MF}}
  \triangleq
  \{ \bm\theta \in \Theta : |\theta_{a,m} - \theta_{a,M}| \le \xi_m, \forall a \in [K], m \in [M] \},
  \]
  where $\xi_m$ is the known bias level for fidelity $m$.
  Assume throughout that $\xi_M=0$.
  The set of possible answers is $\mathcal A = [K]$.
  The goal is to identify the correct answer defined as
  \[
    \Ans(\bm\theta)
    = 
    \begin{cases}
      \argmax_{a \in [K]} \theta_{a,M} & \text{if } \bm\theta \in \Theta_{\mathrm{MF}}, \\
      \varnothing & \text{if } \bm\theta \notin \Theta_{\mathrm{MF}}.
    \end{cases}
  \]
  We assume that there is no tie at the highest fidelity.\footnote{We exclude ties at the highest fidelity because such instances lie on the boundary of identifiability.
  If two alternatives are tied at fidelity $M$, then an arbitrarily small perturbation can make either one uniquely optimal.
  In these boundary cases, the GLR is zero, so exact identification is statistically impossible.
  A standard remedy is to relax the objective and seek an $\varepsilon$-good alternative instead; see Example~\ref{ex:good-alternative}.}
  The answerable set is
  \[
  \Theta_{\mathrm{MF}}^\star
  \triangleq
  \{ \bm\theta \in \Theta : \Ans(\bm\theta) \neq \varnothing \} = \Theta_{\mathrm{MF}} \subsetneq \Theta,
  \]
  i.e., the set of mean tables that satisfy the multi-fidelity constraints.
  The acceptance set for answer $i$ is $\Theta^{\star}_i \triangleq \{ \bm\vartheta \in \Theta^\star_{\mathrm{MF}} : \vartheta_{i,M} > \vartheta_{j,M}, \forall j \neq i \}$.
\end{example}

\begin{example}[Dueling bandits with a Condorcet winner]\label{ex:dueling}
  In dueling bandits, the observations are not direct measurements of individual alternatives, but rather binary preferences between pairs of alternatives.
  The measurement set is the set of unordered pairs $\mathcal U = \{ \{i,j\} : 1 \le i < j \le K \}$, where measurement $u=\{i,j\}$ means that the learner requests a duel between alternatives $i$ and $j$.
  With the convention $i<j$, the observation is Bernoulli with mean $\theta_{i,j}$.
  The ambient parameter space is
  \[
  \Theta_{\mathrm{DB}}
  \triangleq
  \{ \bm\theta = (\theta_{i,j})_{1 \le i,j \le K} \in [0,1]^{K \times K} : \theta_{i,j} = 1 - \theta_{j,i},\ \forall i,j \}.
  \]
  The set of possible answers is $\mathcal A = [K]$.
  The goal is to identify the Condorcet winner
  \[
  \Ans(\bm\theta)
  \triangleq
  \bigl\{ i \in [K] : \theta_{i,j} > 1/2,\ \forall j \neq i \bigr\}.
  \]
  A Condorcet winner need not exist if the preference matrix contains cycles, but it is unique if is exists.\footnote{We can accommodate genuinely non-answerable instances by augmenting the answer set with a null answer $a=\varnothing$.}
  The answerable set is
  \[
  \Theta_{\mathrm{CW}}^\star
  \triangleq
  \{ \bm\theta \in \Theta_{\mathrm{DB}} : |\Ans(\bm\theta)| = 1 \} \subsetneq \Theta_{\mathrm{DB}}.
  \]
  The acceptance set for answer $i$ is
  \[
  \Theta^{\star}_i
  \triangleq
  \{ \bm\vartheta \in \Theta_{\mathrm{CW}}^\star : \vartheta_{i,j} > 1/2,\ \forall j \neq i \}.
  \]
  See also \citet{chen2020combinatorial,haddenhorst2021identification}.
\end{example}

In Examples~\ref{ex:MF-BAI} and~\ref{ex:dueling}, the answerable set $\Theta^\star = \bigcup_{a\in\mathcal A} \Theta^{\star}_a$ is a proper subset of the ambient model $\Theta$.
Consequently, during learning the empirical or posterior estimate may leave $\Theta^\star$, in which case an empirical answer need not exist.
We refer to such estimates as \textit{non-answerable estimates}.

\subsection{Stopping rule via GLR statistics}

We follow the sequential hypothesis-testing perspective in \citet{HongFanLuo2021Review,wang2024bonferroni}.
In the presence of non-answerable estimates, the natural testing statistic is the restricted generalized likelihood ratio (GLR) that compares the best fit inside an acceptance set with the best \textit{answerable parameter} outside.

Let $\widehat{\bm\theta}_n$ denote an unconstrained maximum likelihood estimator (MLE) in the ambient parameter space $\Theta$.
For the one-parameter exponential-family models, the negative log-likelihood ratio relative to $\widehat{\bm\theta}_n$ can be written as a sum of coordinatewise Kullback--Leibler (KL) divergences, i.e., $\Gamma(\bm N_n;\widehat{\bm\theta}_{n}, \bm\vartheta)$, where
\[
  \Gamma(\bm w; \bm \theta, \bm\vartheta) \triangleq \sum_{u \in \mathcal U} w_{u} \mathrm{KL}(P_{\bm\theta,u} \| P_{\bm\vartheta,u}), \quad \text{ for } \bm w \in \R_+^{\mathcal{U}},
\]
where $\mathrm{KL}(\cdot \| \cdot)$ is the KL divergence between probability distributions.
The GLR statistic for answer $a$ is
\begin{equation}\label{eq:GLR}
  Z_n(a) \triangleq Z(a; \bm N_n, \widehat{\bm\theta}_{n}), \quad \text{where} \quad
  Z(a; \bm p, \bm \theta)
  \triangleq
  \inf_{\bm\vartheta \in \mathrm{Alt}_a} \Gamma(\bm p; \bm \theta, \bm\vartheta)
  -
  \inf_{\bm\vartheta \in \Theta^{\star}_a} \Gamma(\bm p; \bm \theta, \bm\vartheta).
\end{equation}
The statistic $Z_n(a)$ is large when the data are much better explained by $\Theta^{\star}_a$ than by any alternative answer.
Let $L_n(\bm\vartheta) \triangleq \Gamma(\bm N_n; \widehat{\bm\theta}_n, \bm\vartheta)$, then $Z_n(a) = \inf_{\bm\vartheta \in \mathrm{Alt}_a} L_n(\bm\vartheta) - \inf_{\bm\vartheta \in \Theta^{\star}_a} L_n(\bm\vartheta)$.

Whenever $\widehat{\bm\theta}_n \in \Theta^{\star}_a$, we have $\inf_{\bm\vartheta \in \Theta^{\star}_a} L_n(\bm\vartheta) = 0$ and hence $Z_n(a) = \inf_{\bm\vartheta \in \mathrm{Alt}_a} L_n(\bm\vartheta)$.
Thus, for a correct answer $a$ and a consistent estimator that eventually enters $\Theta^{\star}_a$, the GLR reduces to the usual Chernoff-type quantity used in fixed-confidence pure exploration \citep{garivier2016optimal}.
In this sense, the GLR is a robustified version of the usual Chernoff information to handle non-answerable estimates.

Recall that $Z_n(a)$ quantifies the evidence for $a$ to be correct.
The overall evidence of the data for \textit{some} answer to be correct is then $\max_{a \in \mathcal A} Z_n(a)$.
Accordingly, a GLR stopping rule is
\begin{equation}\label{eq:GLR-stopping}
  \tau_\delta \triangleq \inf \{ n \ge 1 : \max_{a \in \mathcal A} Z_n(a) \ge \beta_{n,\delta} \},
\end{equation}
where $\beta_{n,\delta}$ is a threshold typically in the order of $\mathcal{O}(\log(1 / \delta))$; see \citet{ding2025stopping} for discussions.
The corresponding decision rule is to nominate the answer best-supported by the data:
\begin{equation}\label{eq:decision}
  \hat a_{\tau_\delta} \in \argmax_{a \in \mathcal A} Z_{\tau_\delta}(a).
\end{equation}
This decision rule differs from the usual empirical-best-answer rule.
We choose it for essentially free finite-sample robustness, since the noisy estimates may be non-answerable and the empirical-best answer may then be undefined.
Such transient non-answerability vanishes asymptotically under accurate estimation.

\begin{remark}[On the threshold $\beta_{n,\delta}$ and $\delta$-correctness]
  One needs a threshold sequence $\beta_{n,\delta}$ such that
  $
  \Prob_{\bm\theta}\left(
  \exists a \notin \Ans(\bm\theta),\ \exists n \ge 1 : Z_n(a) \ge \beta_{n,\delta}
  \right) \le \delta
  $
  for every $\bm\theta \in \Theta^\star$.
  This is the usual generalized-likelihood-ratio logic behind fixed-confidence pure exploration.
  The exact threshold calibration is generally open and problem-dependent; see \citet{garivier2016optimal,ramdas2020admissible}, and \citet{ding2025stopping} for related discussions from the machine learning, economics and simulation communities.
  For the purposes of this paper, we keep the exact threshold abstract and focus instead on efficient sampling rules.
  In the numerical experiments, we use the heuristic threshold $\beta_{n,\delta} = \log\bigl((\log n + 1)/\delta\bigr)$.
\end{remark}

\subsection{Answer-wise pitfalls and the max-max-min characteristic value}

To state the lower bound in a way that covers both non-answerable estimates and multiple-correct-answer problems, we propose an answer-wise generalization of pitfalls in \citet{qin2025dual}.
For each potential answer $a \in \mathcal A$, we assume that the rejection region $\mathrm{Alt}_a$ can be decomposed into finitely many simpler alternative sets.
More precisely, for every candidate answer $a \in \mathcal A$, let
\[
  \mathrm{Alt}_a = \bigcup_{x \in \X_a} \mathrm{Alt}_{a,x},
\]
where $\X_a$ is a finite pitfall index set for answer $a$ and each $\mathrm{Alt}_{a,x}\subseteq\mathrm{Alt}_a$ is a convex set of (answerable) alternative instances associated with pitfall $x$ for answer $a$; see Sections~\ref{sec:good_alternative_formula}--\ref{sec:dueling}) for examples.
A pitfall $x\in\X_a$ describes one concrete way in which answer $a$ can fail.
In ranking-and-selection, a pitfall is a challenger alternative that beats the current nominee.
In dueling bandits, a pitfall is an opponent that prevents the nominee from being a Condorcet winner.

In \citet{qin2025dual}, pitfalls are indexed through the unique correct answer at the true instance.
Here we instead index pitfalls by each candidate answer $a$.
This answer-wise indexing is what lets the framework handle both multiple correct answers and temporary non-answerable estimates.

For lower-bound statements, we consider only answerable instances  $\bm\theta \in \Theta^\star$.
For any allocation $\bm p \in \Delta_{\mathcal U}$, true instance $\bm\theta \in \Theta^\star$, and alternative instance $\bm\vartheta \in \Theta$, define the cost-normalized discrimination rate
\[
  \Gamma^{\mathrm c}(\bm p;\bm\theta,\bm\vartheta)
  \triangleq
  \frac{\sum_{u \in \mathcal U} p_u \mathrm{KL}(P_{\bm\theta,u} \| P_{\bm\vartheta,u})}
  {\bar C_{\bm\theta}(\bm p)}.
\]
This is the statistical distance between $\bm\theta$ and $\bm\vartheta$ under allocation $\bm p$, normalized by the average cost per sample.
For each answer--pitfall pair $(a,x)$, define the cost-normalized information rate for ruling out pitfall $x$ when certifying answer $a$ as
\[
  D_{a,x}(\bm p;\bm\theta)
  \triangleq
  \inf_{\bm\vartheta \in \mathrm{Alt}_{a,x}}
  \Gamma^{\mathrm c}(\bm p;\bm\theta,\bm\vartheta).
\]
Let $\bm\vartheta^{a,x}(\bm p, \bm\theta) \in \argmin_{\bm\vartheta \in \overline{\mathrm{Alt}}_{a,x}}
\Gamma^{\mathrm c}(\bm p;\bm\theta,\bm\vartheta)$ denote the most-confusing alternative instance.

The learner must rule out \emph{all} pitfalls, so the bottleneck information rate is $\min_{x \in \X_a} D_{a,x}(\bm p;\bm\theta)$.
Among all potential answers in $\mathcal A$, the learner may nominate the one that is easiest to certify, namely the one with the largest bottleneck information rate.
The learner then chooses the allocation $\bm p$ to maximize that rate.
This leads to the characteristic value
\[
  \Gamma^\star(\bm\theta)
  \triangleq
  \max_{\bm p \in \Delta_{\mathcal U}}
  \max_{a \in \mathcal{A}}
  \min_{x \in \X_a}
  D_{a,x}(\bm p;\bm\theta).
\]
For problems with a unique answer, the second maximization is vacuous, as it is always attained by that answer.
Accordingly, the characteristic value reduces to the one studied in \citet{qin2025dual}.
Nevertheless, the second maximization remains important when temporary estimates fall outside the answerable set.
In multiple-correct-answer problems \citep{degenne2019pure}, it is essential even for answerable instances, because the answer set is set-valued.

Under any allocation rule that guarantees consistent estimation of the parameters, one expects $Z_n(a)$ to grow approximately at rate
$B_n \cdot \min_{x \in \X_a} D_{a,x}(\bm p;\bm\theta).$
Optimizing over $\bm p$ and $a$ suggests the heuristic scaling
$
\max_{a \in \mathcal A} Z_n(a) \approx B_n \cdot \Gamma^\star(\bm\theta).
$
In view of the stopping rule \eqref{eq:GLR-stopping}, this leads to the heuristic lower bound:
\[
  \E_{\bm\theta}[B_{\tau_\delta}] \gtrsim \frac{\log(1/\delta)}{\Gamma^\star(\bm\theta)}.
\]
This is the direct analogue of the classical characteristic time in fixed-confidence pure exploration \citep{garivier2016optimal}, except that the normalization is by spent budget rather than by the number of samples.
We refer to \citet{qin2024optimizing} for a rigorous discussion.

\section{The ENDS algorithm}\label{sec:algorithm}

We now generalize \citet{qin2025dual} to accommodate multiple correct answers and non-answerable estimates.
At round $n$, let $\bm\theta_n \in \Theta$ be an estimator of $\bm\theta$.
This estimator may be the sample mean, a posterior draw, or a projection onto the answerable set.
During sampling, we nominate
\[
  a_n \in \arg\max_{a\in\mathcal A} Z(a; \bm p_n, \bm \theta_n)
\]
as the most plausible answer.
(Here, we use a variant of GLR, namely $Z(a; \bm p_n, \bm \theta_n)$, which uses the user-specified estimator $\bm\theta_n$
instead of the ambient MLE $\widehat{\bm\theta}_n$.
In our numerical experiments, we use posterior sampling as the default estimation routine for the sampling rule.
Drawing $\bm\theta_n$ from the current posterior naturally encourages exploration and mitigates the influence of rare, high-leverage early observations on nomination and pitfall detection.)
Once the nominee $a_n$ has been selected, the pitfall detector chooses
\[
  x_n \in \argmin_{x \in \X_{a_n}} D_{a_n,x}(\bm p_n;\bm\theta_n).
\]
The principal pitfall $x_n$ is the most likely way for answer $a_n$ to fail under the current estimate and allocation.
The cost-aware information-directed selection (IDS) distribution is then
\begin{equation}\label{eq:IDS}
  H_u^{a_n,x_n}(\bm p_n;\bm\theta_n)
  \triangleq
  \frac{p_{n,u} \mathrm{KL}(P_{\bm\theta_n,u} \| P_{\bm\vartheta^{a_n,x_n},u}) / C_u(\bm\theta_n)}{
  \sum_{v \in \mathcal U} p_{n,v} \mathrm{KL}(P_{\bm\theta_n,v} \| P_{\bm\vartheta^{a_n,x_n},v}) / C_v(\bm\theta_n)},
\end{equation}
where $\bm\vartheta^{a_n,x_n} = \bm\vartheta^{a_n,x_n}(\bm p_n, \bm\theta_n)$ is the most confusing alternative instance for that pitfall, namely
\[
  \bm\vartheta^{a_n,x_n}(\bm p_n, \bm\theta_n)
  \in
  \argmin_{\bm\vartheta \in \overline{\mathrm{Alt}}_{a_n,x_n}}
  \Gamma^{\mathrm c}(\bm p_n;\bm\theta_n,\bm\vartheta).
\]
The IDS \eqref{eq:IDS} defines a valid distribution over measurement types; see \citet{qin2025dual} for the case with unit-cost.
The next measurement is drawn from this distribution.
The stopping rule is given by \eqref{eq:GLR-stopping} and final decision is the nominee in \eqref{eq:decision}; both based on the ambient-MLE-based GLR \eqref{eq:GLR}.
This yields a conceptually simple four-step scheme: Estimate--Nominate--Detect--Select.
We refer to this algorithm as ENDS and summarize its fixed-precision variant in Algorithm~\ref{alg:generalized_pan}.
The sampling rule is anytime, so the algorithm can be turned into a fixed-budget algorithm by replacing the stopping rule with a fixed budget.

\begin{algorithm}[hbtp]
  \caption{ENDS}
  \label{alg:generalized_pan}
  \begin{algorithmic}[1]
    \Require Confidence level $\delta$, threshold sequence $\{\beta_{n,\delta}\}_{n\ge1}$, estimation routine \name{est} (default: posterior sampling)
    \State \textbf{Initialize:} sample each measurement type $u\in\mathcal U$ once
    \While{$\max_{a\in\mathcal A} Z_n(a) < \beta_{n,\delta}$}
      \State $\bm\theta_n \gets \name{est}(\mathcal H_n)$ \Comment{Estimate}
      \State $a_n \in \arg\max_{a\in\mathcal A} Z(a; \bm p_n, \bm \theta_n)$ \Comment{Nominate}
      \State $x_n \in \arg\min_{x\in\mathcal X_{a_n}} D_{a_n,x}(\bm p_n;\bm\theta_n)$ \Comment{Detect}
      \State draw $U_{n+1}\sim \bm H^{a_n,x_n}(\bm p_n;\bm\theta_n)$, where $\bm H^{a_n,x_n} = (H_u^{a_n,x_n})_{u\in\mathcal U}$ is defined in \eqref{eq:IDS} \Comment{Select}
      \State measure $U_{n+1}$, observe $Y_{n+1}$ and update $\mathcal H_{n+1}$
    \EndWhile
    \State \Return $\hat a_n \in \arg\max_{a\in\mathcal A} Z_n(a)$
  \end{algorithmic}
\end{algorithm}

To instantiate this algorithm for a specific problem, one needs only to specify the explicit form of the alternative sets $\mathrm{Alt}_{a,x}$, the cost-normalized information rates $D_{a,x}(\bm p;\bm\theta)$, and the IDS distribution $\bm H^{a,x}(\bm p;\bm\theta)$.
We now provide these explicit forms for Examples~\ref{ex:good-alternative}--\ref{ex:dueling}.

\subsection{Example: Selection of a good alternative}\label{sec:good_alternative_formula}
Consider the selection of a good alternative in Example~\ref{ex:good-alternative}.
A natural pitfall for answer $a$ is a challenger alternative $x\neq a$ whose mean exceeds that of $a$ by more than $\varepsilon$.
Accordingly, for $x \in [K]\setminus\{a\}$ define
$
\mathrm{Alt}_{a,x}^{\varepsilon}
\triangleq
\{ \bm\vartheta \in \Theta : \vartheta_x > \vartheta_a + \varepsilon \}.
$
Then $\mathrm{Alt}_a = \bigcup_{x\neq a} \mathrm{Alt}_{a,x}^{\varepsilon}$ and $\X_a=[K]\setminus\{a\}$.
The characteristic value is
\[
  \Gamma_{\varepsilon}^\star(\bm\theta)
  =
  \max_{\bm p \in \Delta_K}
  \max_{a \in [K]}
  \min_{x \neq a}
  D_{a,x}^{\varepsilon}(\bm p;\bm\theta),
  \quad\text{where}\quad
  D_{a,x}^{\varepsilon}(\bm p;\bm\theta)
  \triangleq
  \inf_{\bm\vartheta \in \mathrm{Alt}_{a,x}^{\varepsilon}}
  \Gamma^{\mathrm c}(\bm p;\bm\theta,\bm\vartheta).
\]

Assume that all alternatives belong to the same one-parameter exponential family with mean domain $\mathsf M$ and KL divergence $d$.
For $\mu\in\mathsf M$, define the upper-feasible interval
$
J_\varepsilon(\mu)\triangleq (-\infty,\mu+\varepsilon]\cap \mathsf M.
$
For $z\in\mathsf M$, let $\Pi_{J_\varepsilon(\mu)}(z)$ denote the projection of $z$ onto $J_\varepsilon(\mu)$.
For any weight vector $\bm w=(w_i)_{i\in[K]}\in\R_+^K$ and any mean vector $\bm z=(z_i)_{i\in[K]}\in \mathsf M^K$, define
\[
  G_{a,x}^{\varepsilon}(\bm w,\bm z)
  \triangleq
  \inf_{\substack{\mu,\nu\in\mathsf M\\ \nu\ge \mu+\varepsilon}}
  \Bigl\{
  w_a  d(z_a,\mu) + w_x  d(z_x,\nu)
  \Bigr\}
\]
and
\[
  A_a^{\varepsilon}(\bm w,\bm z)
  \triangleq
  \inf_{\mu\in\mathsf M}
  \left[
  w_a  d(z_a,\mu)
  +
  \sum_{j\neq a}
  w_j  d\left(z_j,\Pi_{J_\varepsilon(\mu)}(z_j)\right)
  \right].
\]
The quantity $G_{a,x}^{\varepsilon}(\bm w,\bm z)$ is the weighted KL distance from $\bm z$ to the pitfall set $\mathrm{Alt}_{a,x}^{\varepsilon}$, while $A_a^{\varepsilon}(\bm w,\bm z)$ is the weighted KL distance from $\bm z$ to the acceptance set $\Theta_a^\star$.
Then,
$
D_{a,x}^{\varepsilon}(\bm p;\bm\theta)
=
G_{a,x}^{\varepsilon}(\bm p,\bm\theta) / \bar C_{\bm\theta}(\bm p).
$
Let
$
(\mu_{a,x}^\star,\nu_{a,x}^\star)
\in
\arg\min_{\mu,\nu\in\mathsf M: \nu\ge \mu+\varepsilon}
\bigl\{
p_a  d(\theta_a,\mu)+p_x  d(\theta_x,\nu)
\bigr\},
$
and define $\vartheta_i^{a,x}(\bm p,\bm\theta) =
\mu_{a,x}^\star$ if $i=a$, $\nu_{a,x}^\star$ if $i=x$, and $\theta_i$ otherwise.
\begin{proposition}\label{prop:good-alternative-explicit}
  The vector $\bm\vartheta^{a,x}(\bm p,\bm\theta)$ attains the value $G_{a,x}^{\varepsilon}(\bm p,\bm\theta)$.
  Plugging $\bm{\vartheta}^{a,x}$ into \eqref{eq:IDS}, we obtain the explicit IDS distribution $\bm{H}^{a,x}(\bm p;\bm\theta)$.
  Specifically, $H_i^{a,x}(\bm p;\bm\theta)=0$ for every $i\notin\{a,x\}$.
  Let $\bm N_n=(N_{n,i})_{i\in[K]}$ and let $\widehat{\bm\theta}_n$ be the ambient MLE.
  The explicit form of $Z_n(a)$ is obtained by
  \[
    \inf_{\bm\vartheta \in \Theta_a^\star} L_n(\bm\vartheta)
    =
    A_a^{\varepsilon}(\bm N_n,\widehat{\bm\theta}_n),
    \qquad
    \inf_{\bm\vartheta \in \mathrm{Alt}_a} L_n(\bm\vartheta)
    =
    \min_{x\neq a} G_{a,x}^{\varepsilon}(\bm N_n,\widehat{\bm\theta}_n).
  \]
\end{proposition}

\begin{remark}
  In the Gaussian unit-variance, unit-cost case, $d(u,v)=(u-v)^2/2$ and $\mathsf M=\R$.
  If $\theta_x < \theta_a+\varepsilon$, then
  $
  G_{a,x}^{\varepsilon}(\bm p,\bm\theta)
  =
  \frac{(\theta_a-\theta_x+\varepsilon)^2}{2\left(1/p_a+1/p_x\right)}.
  $
  The optimizer is
  $
  \mu_{a,x}^\star
  =
  \frac{p_a\theta_a+p_x(\theta_x-\varepsilon)}{p_a+p_x}$ and $\nu_{a,x}^\star = \mu_{a,x}^\star+\varepsilon$,
  and the IDS rule reduces to $H_a^{a,x}(\bm p;\bm\theta) = \frac{p_x}{p_a+p_x}$ and $H_x^{a,x}(\bm p;\bm\theta) = \frac{p_a}{p_a+p_x}$.
  The outer maximization is attained by $a=I^\star$, and
  \[
    \Gamma_{\varepsilon}^\star(\bm\theta)
    =
    \max_{\bm p\in\Delta_K}
    \min_{x\neq I^\star}
    \frac{(\theta_{I^\star}-\theta_x+\varepsilon)^2}{2\left(1/p_{I^\star}+1/p_x\right)}.
  \]
\end{remark}

\subsection{Example: Multi-fidelity ranking-and-selection}
Consider the multi-fidelity ranking-and-selection problem described in Example~\ref{ex:MF-BAI}.
A natural pitfall for answer $a$ is a challenger alternative $x\neq a$ whose top-fidelity mean strictly exceeds that of alternative $a$.
Accordingly, for $x \in [K]\setminus\{a\}$, define
$
\mathrm{Alt}^{\mathrm{MF}}_{a,x}
\triangleq
\{ \bm\vartheta \in \Theta_{\mathrm{MF}} : \vartheta_{x,M} > \vartheta_{a,M} \}.
$
The characteristic value is
\begin{equation}\label{eq:Gamma_MF}
  \Gamma_{\mathrm{MF}}^\star(\bm\theta)
  =
  \max_{\bm p \in \Delta_{\mathcal U}}
  \max_{a \in \mathcal A}
  \min_{x \neq a}
  D^{\mathrm{MF}}_{a,x}(\bm p;\bm\theta),
  \quad\text{where}\quad
  D^{\mathrm{MF}}_{a,x}(\bm p;\bm\theta)
  \triangleq
  \inf_{\bm\vartheta \in \mathrm{Alt}^{\mathrm{MF}}_{a,x}}
  \Gamma^{\mathrm c}(\bm p;\bm\theta,\bm\vartheta).
\end{equation}
At answerable instances with a unique best alternative, the maximum over $a$ is vacuous because it is attained by that alternative.
For non-answerable instances, the maximization is useful for nomination.

Assume that each fidelity level belongs to a one-parameter exponential family with mean domain $\mathsf M$ and KL divergence $d$.
For $\mu\in\mathsf M$, define the valid interval at fidelity $m$ by $I_m(\mu)\triangleq [\mu-\xi_m,\mu+\xi_m]\cap \mathsf M$ for all $m\in[M]$.
For $z\in\mathsf M$, let $\Pi_{I_m(\mu)}(z)$ denote the projection of $z$ onto $I_m(\mu)$.
For any weight $\bm w=(w_{i,m})_{i\in[K], m\in[M]}$ and any mean table
$\bm z=(z_{i,m})_{i\in[K], m\in[M]}\in \mathsf M^{K\times M}$, define
$
\Psi_i(\mu;\bm w,\bm z)
\triangleq
\sum_{m=1}^M w_{i,m}
d\left(z_{i,m}, \Pi_{I_m(\mu)}(z_{i,m})\right)
$ for $i\in[K]$.
This is the weighted KL distance from $\bm z$ to the set of mean tables that satisfy the multi-fidelity constraints for alternative $i$ with top-fidelity mean $\mu$.

For $a\in[K]$ and $x\neq a$, define the weighted KL distance from $\bm z$ to the pitfall set $\mathrm{Alt}_{a,x}^{\mathrm{MF}}$
\begin{align*}
  G_{a,x}(\bm w,\bm z)
  & \triangleq
  \inf_{\bm\vartheta \in \mathrm{Alt}^{\mathrm{MF}}_{a,x}}
  \sum_{\substack{i\in[K]\\ m\in[M]}} w_{i,m}  d(z_{i,m},\vartheta_{i,m}) \\
  & =
  \sum_{i\notin\{a,x\}} \inf_{\mu\in\mathsf M}\Psi_i(\mu;\bm w,\bm z)
  +
  \inf_{\substack{\mu_a,\mu_x\in\mathsf M\\\mu_x\ge \mu_a}}
  \Bigl\{
  \Psi_a(\mu_a;\bm w,\bm z)+\Psi_x(\mu_x;\bm w,\bm z)
  \Bigr\}.
\end{align*}

For each $j\neq a$, fix any minimizer $\ell_j\in \arg\min_{\nu\in\mathsf M}\Psi_j(\nu;\bm w,\bm z)$,
and define
\[
  A_a(\bm w,\bm z)
  \triangleq
  \inf_{\mu\in\mathsf M}
  \Bigl[
  \Psi_a(\mu;\bm w,\bm z)
  +
  \sum_{j\neq a}
  \Psi_j(\mu\wedge \ell_j;\bm w,\bm z)
  \Bigr].
\]
Since each $\Psi_j(\cdot;\bm w,\bm z)$ is convex, $\inf_{\nu\le \mu}\Psi_j(\nu;\bm w,\bm z)=\Psi_j(\mu\wedge \ell_j;\bm w,\bm z)$ for all $j\neq a$,
so $A_a(\bm w,\bm z)$ is exactly the weighted KL distance from $\bm z$ to the acceptance set $\Theta_a^\star$, written as a one-dimensional convex optimization.
Finally, $D^{\mathrm{MF}}_{a,x}(\bm p;\bm\theta)=G_{a,x}(\bm p,\bm\theta) / \bar C(\bm p) $ with $\bar C(\bm p)\triangleq \sum_{i=1}^K\sum_{m=1}^M p_{i,m} C_m$.

Let $\mu_i^\star\in\arg\min_{\mu\in\mathsf M}\Psi_i(\mu;\bm p,\bm\theta)$ for each $i\notin\{a,x\}$, let
\[
  (\mu_a^\star,\mu_x^\star)\in
  \arg\min_{\mu_a,\mu_x\in\mathsf M: \mu_x\ge \mu_a}
  \bigl\{
  \Psi_a(\mu_a;\bm p,\bm\theta)+\Psi_x(\mu_x;\bm p,\bm\theta)
  \bigr\},
\]
fix $\ell_j^\star\in\arg\min_{\nu\in\mathsf M}\Psi_j(\nu;\bm p,\bm\theta)$ for each $j\neq a$, and let
\[
  \lambda_a^\star
  \in
  \arg\min_{\mu\in\mathsf M}
  \bigl[
  \Psi_a(\mu;\bm p,\bm\theta)
  +
  \sum_{j\neq a}
  \Psi_j(\mu\wedge \ell_j^\star;\bm p,\bm\theta)
  \bigr].
\]
\begin{proposition}\label{prop:MF-explicit}
  The table $\bm\vartheta^{a,x}(\bm p,\bm\theta)$ defined by $\vartheta_{i,m}^{a,x}(\bm p,\bm\theta)
  \triangleq
  \Pi_{I_m(\mu_i^\star)}(\theta_{i,m})$ for $i\in[K],\ m\in[M]$
  attains the value $G_{a,x}(\bm p,\bm\theta)$.
  Plugging $\bm{\vartheta}^{a,x}$ into \eqref{eq:IDS}, we obtain the explicit IDS distribution $\bm{H}^{a,x}(\bm p;\bm\theta)$.
  The table $\bm\vartheta^{a,\star}(\bm p,\bm\theta)$ defined below
  belongs to $\Theta_a^\star$ and attains the value $A_a(\bm p,\bm\theta)$:
  \[
    \vartheta_{i,m}^{a,\star}(\bm p,\bm\theta)
    \triangleq
    \begin{cases}
      \Pi_{I_m(\lambda_a^\star)}(\theta_{a,m}), & i=a,\\
      \Pi_{I_m(\lambda_a^\star\wedge \ell_i^\star)}(\theta_{i,m}), & i\neq a,
    \end{cases}
    \qquad i\in[K],\ m\in[M].
  \]
  Let $\bm N_n=(N_{n,i,m})_{i,m}$ and let $\widehat{\bm\theta}_n$ be the ambient MLE.
  The explicit form of $Z_n(a)$ is obtained by
  \[
    \inf_{\bm\vartheta\in\Theta_a^\star} L_n(\bm\vartheta)
    =
    A_a(\bm N_n,\widehat{\bm\theta}_n),
    \qquad
    \inf_{\bm\vartheta\in\mathrm{Alt}_a} L_n(\bm\vartheta)
    =
    \min_{x\neq a} G_{a,x}(\bm N_n,\widehat{\bm\theta}_n).
  \]
\end{proposition}

\begin{remark}[Gaussian special case]
  In the Gaussian unit-variance case, $\mathsf M=\R$ and $d(u,v)=(u-v)^2/2$, so
  \[
    \Psi_i(\mu;\bm w,\bm z)
    =
    \frac12\sum_{m=1}^M w_{i,m}\bigl(|z_{i,m}-\mu|-\xi_m\bigr)_+^2,
    \qquad i\in[K].
  \]
  Therefore, for any choice of minimizers $\ell_j\in\arg\min_{\nu\in\R}\Psi_j(\nu;\bm w,\bm z)$,
  \[
    A_a(\bm w,\bm z)
    =
    \inf_{\mu\in\R}
    \left[
    \frac12\sum_{m=1}^M w_{a,m}\bigl(|z_{a,m}-\mu|-\xi_m\bigr)_+^2
    +
    \sum_{j\neq a}\frac12\sum_{m=1}^M
    w_{j,m}\bigl(|z_{j,m}-(\mu\wedge \ell_j)|-\xi_m\bigr)_+^2
    \right].
  \]
\end{remark}

\begin{remark}
  If $\widehat{\bm\theta}_n\in\Theta_a^\star$, then $A_a(\bm N_n,\widehat{\bm\theta}_n)=0$ and, for every $x\neq a$,
  \[
    G_{a,x}(\bm N_n,\widehat{\bm\theta}_n)
    =
    \inf_{\mu\in\mathsf M}
    \bigl\{
    \Psi_a(\mu;\bm N_n,\widehat{\bm\theta}_n)
    +
    \Psi_x(\mu;\bm N_n,\widehat{\bm\theta}_n)
    \bigr\}.
  \]
  Hence
  $
  Z_n(a)
  =
  \min_{x\neq a}\inf_{\mu\in\mathsf M}
  \bigl\{
  \Psi_a(\mu;\bm N_n,\widehat{\bm\theta}_n)
  +
  \Psi_x(\mu;\bm N_n,\widehat{\bm\theta}_n)
  \bigr\}.
  $
\end{remark}

\subsection{Example: Dueling bandits with a Condorcet winner}\label{sec:dueling}
Consider the dueling bandit problem described in Example~\ref{ex:dueling}.
A natural pitfall for answer $a$ is a challenger alternative $x\neq a$ that itself becomes a Condorcet winner.
Accordingly, for $x\in[K]\setminus\{a\}$, define
$
\mathrm{Alt}^{\mathrm{CW}}_{a,x}
\triangleq
\{ \bm\vartheta \in \Theta_{\mathrm{DB}} : \vartheta_{x,j} > 1/2,\ \forall j \neq x \}.
$
Since $x$ beating every opponent already implies uniqueness of the Condorcet winner, we have
$
\mathrm{Alt}^{\mathrm{CW}}_{a,x}=\Theta_x^\star
$.
Hence
$
\mathrm{Alt}_a
=
\bigcup_{x\neq a}\mathrm{Alt}^{\mathrm{CW}}_{a,x}
$
with $\X_a=[K]\setminus\{a\}$.
The characteristic value is
\[
  \Gamma_{\mathrm{CW}}^\star(\bm\theta)
  =
  \max_{\bm p \in \Delta_{\mathcal U}}
  \max_{a \in [K]}
  \min_{x \neq a}
  D_{a,x}^{\mathrm{CW}}(\bm p;\bm\theta),
  \quad\text{where}\quad
  D_{a,x}^{\mathrm{CW}}(\bm p;\bm\theta)
  \triangleq
  \inf_{\bm\vartheta \in \mathrm{Alt}^{\mathrm{CW}}_{a,x}}
  \Gamma^{\mathrm c}(\bm p;\bm\theta,\bm\vartheta).
\]

Let $d$ denote the Bernoulli KL divergence.
For any weight vector
$\bm w=(w_{\{i,j\}})_{\{i,j\}\in\mathcal U}\in\R_+^{\mathcal U}$,
extend it symmetrically by setting
$w_{i,j}\triangleq w_{\{i,j\}}=w_{j,i}$ for $i\neq j$.
Define $I_{\mathrm{CW}}\triangleq [1/2,1]$ and projection $\Pi_{\mathrm{CW}}(q)\triangleq \Pi_{I_{\mathrm{CW}}}(q)=q\vee \frac12$ for $q\in[0,1]$.
For any preference matrix $\bm z\in \Theta_{\mathrm{DB}}$ and each alternative $i\in[K]$, define
\[
  \Psi_i(\bm w,\bm z)
  \triangleq
  \sum_{j\neq i}
  w_{i,j}
  d\left(z_{i,j},\Pi_{\mathrm{CW}}(z_{i,j})\right)
  =
  \sum_{j:z_{i,j}\le 1/2}
  w_{i,j}  d(z_{i,j},1/2).
\]
Consequently,
$
G_{a,x}^{\mathrm{CW}}(\bm w,\bm z)
=
\Psi_x(\bm w,\bm z),
$
$
A_a^{\mathrm{CW}}(\bm w,\bm z)
=
\Psi_a(\bm w,\bm z),
$
and
\[
  D_{a,x}^{\mathrm{CW}}(\bm p;\bm\theta)
  =
  \frac{\Psi_x(\bm p,\bm\theta)}{\bar C_{\bm\theta}(\bm p)}
  =
  \frac{\sum_{j\neq x} p_{\{x,j\}}  d(\theta_{x,j},\theta_{x,j}\vee 1/2)}
  {\bar C_{\bm\theta}(\bm p)}
  =
  \frac{\sum_{j:\theta_{x,j}\le 1/2} p_{\{x,j\}}  d(\theta_{x,j},1/2)}
  {\bar C_{\bm\theta}(\bm p)}.
\]

The most confusing alternative for pitfall $x$ is therefore obtained by leaving every duel not involving $x$ unchanged and pushing each losing duel of $x$ exactly to the boundary $1/2$.
For $x\neq a$, define the boundary representative
$\bm\vartheta^{a,x}(\bm p,\bm\theta)\in\Theta_{\mathrm{DB}}$ by
$
\vartheta^{a,x}_{x,j}(\bm p,\bm\theta)
\triangleq
\theta_{x,j}\vee \frac12$,
$
\vartheta^{a,x}_{j,x}(\bm p,\bm\theta)
\triangleq
1-\vartheta^{a,x}_{x,j}(\bm p,\bm\theta), j\neq x,
$
and
$
\vartheta^{a,x}_{i,j}(\bm p,\bm\theta)
\triangleq
\theta_{i,j}
$
for all $i,j\neq x$.
Similarly, define $\bm\vartheta^{a,\star}(\bm p,\bm\theta)$ by replacing $x$ with $a$ in the display above.

\begin{proposition}\label{prop:dueling-explicit}
  The matrix $\bm\vartheta^{a,x}(\bm p,\bm\theta)$ attains the same objective value as the infimum defining
  $G_{a,x}^{\mathrm{CW}}(\bm p,\bm\theta)$, and
  $\bm\vartheta^{a,\star}(\bm p,\bm\theta)$ attains the same objective value as the infimum defining
  $A_a^{\mathrm{CW}}(\bm p,\bm\theta)$.
  Plugging $\bm\vartheta^{a,x}$ into \eqref{eq:IDS}, we obtain the explicit IDS distribution
  $\bm H^{a,x}(\bm p;\bm\theta)$.
  Let $\bm N_n=(N_{n,\{i,j\}})_{\{i,j\}\in\mathcal U}$ and let $\widehat{\bm\theta}_n$ be the ambient MLE.
  The explicit form of $Z_n(a)$ is obtained by
  \[
    \inf_{\bm\vartheta \in \Theta_a^\star} L_n(\bm\vartheta)
    =
    A_a^{\mathrm{CW}}(\bm N_n,\widehat{\bm\theta}_n)
    =
    \Psi_a(\bm N_n,\widehat{\bm\theta}_n)
  \]
  and
  \[
    \inf_{\bm\vartheta \in \mathrm{Alt}_a} L_n(\bm\vartheta)
    =
    \min_{x\neq a}
    G_{a,x}^{\mathrm{CW}}(\bm N_n,\widehat{\bm\theta}_n)
    =
    \min_{x\neq a}
    \Psi_x(\bm N_n,\widehat{\bm\theta}_n).
  \]
\end{proposition}

\begin{remark}
  If $\widehat{\bm\theta}_n \in \Theta_a^\star$, then $\Psi_a(\bm N_n,\widehat{\bm\theta}_n)=0$ and
  \[
    Z_n(a)
    =
    \min_{x\neq a}
    \sum_{j:\widehat{\theta}_{n,x,j}\le 1/2}
    N_{n,\{x,j\}}  d(\widehat{\theta}_{n,x,j},1/2).
  \]
  If $\bm\theta\in\Theta_{\mathrm{CW}}^\star$ has Condorcet winner $a^\star$, then the outer maximization in $\Gamma_{\mathrm{CW}}^\star(\bm\theta)$ is attained by $a^\star$.
  In the unit-cost setting,
  \[
    \Gamma_{\mathrm{CW}}^\star(\bm\theta)
    =
    \max_{\bm p\in\Delta_{\mathcal U}}
    \min_{x\neq a^\star}
    \sum_{j:\theta_{x,j}\le 1/2}
    p_{\{x,j\}}  d(\theta_{x,j},1/2).
  \]
\end{remark}

\section{Numerical experiments}\label{sec:numerical}
In this section, we evaluate the proposed algorithm ENDS against state-of-the-art methods for selection of a good alternative, multi-fidelity R\&S, and dueling bandits.
Our numerical results show that ENDS (with posterior sampling as the default estimation routine), as a unified algorithm, is highly robust and competitive.
We refer readers to \citet{qin2025dual} for more experiments on conventional R\&S, whose algorithm is a special case of ENDS with unique answers.

\subsection{Selection of a good alternative}\label{sec:num_good_alternative}
For Example~\ref{ex:good-alternative}, we evaluate ENDS on $100$ randomly generated Gaussian instances following \citet{zhong2018fully}.
In each instance, the means of the $K$ arms are drawn i.i.d. from $\mathcal{N}(0,16\epsilon^2)$ with $\epsilon =0.1$, and the observation variances are all equal to $1$ but treated as unknown.
As baselines, we implement the unknown-variance version of Algorithm~2 of \citet{kim2024rate} (coupled with our GLR stopping), KN procedure \citep{kim2001fully}, Paulson's procedure \citep{paulson1964sequential}, Procedure~2 of \citet{zhong2018fully}, and uniform allocation.
For ENDS, we plug the empirical sample variances into the formulas in Section~\ref{sec:good_alternative_formula}.
A fully rigorous treatment of unknown variance would require extending the GLR to two-parameter exponential-family models; see Section~4.1 of \citet{qin2025dual}.
ENDS and the uniform-allocation baseline use the stopping rule in \eqref{eq:GLR-stopping} with the heuristic threshold $\beta_{n,\delta} = \log\bigl((\log n + 1)/\delta\bigr)$.

\begin{table}[H]
  \centering
  \caption{Comparison of expected stopping budget under confidence level $\delta=0.05$ and unknown observation variance $\sigma_i^2=1$. PCS omitted, since all algorithms achieve a perfect PCS of 1.}
  \label{tab:eBAI_comparison}
  \footnotesize
  \setlength{\tabcolsep}{3.5pt}
  \renewcommand{\arraystretch}{1.05}
  \begin{tabular}{l c c c c c c c}
    \toprule
    \multirow{2}{*}{\textbf{$K$}} &
    \multirow{2}{*}{\shortstack{\textbf{\# good}\\\textbf{systems}}} &
    \textbf{ENDS} &
    \textbf{KE-Alg2} &
    \textbf{KN} &
    \textbf{Paulson} &
    \textbf{ZH-Proc2} &
    \textbf{Uniform} \\
    \cmidrule(lr){3-8}
    & & $\mathbb{E}[B_{\tau_\delta}] \pm \mathrm{SE}$ &
    $\mathbb{E}[B_{\tau_\delta}] \pm \mathrm{SE}$ &
    $\mathbb{E}[B_{\tau_\delta}] \pm \mathrm{SE}$ &
    $\mathbb{E}[B_{\tau_\delta}] \pm \mathrm{SE}$ &
    $\mathbb{E}[B_{\tau_\delta}] \pm \mathrm{SE}$ &
    $\mathbb{E}[B_{\tau_\delta}] \pm \mathrm{SE}$ \\
    \midrule
    10   & 1.43 & $\mathbf{1035}\pm122$  & $1723\pm305$  & $2545\pm159$    & $3236\pm199$    & $6243\pm563$    & $3737\pm414$ \\
    100  & 1.86 & $\mathbf{4563}\pm279$  & $9148\pm1702$ & $21453\pm482$   & $27707\pm701$   & $42028\pm1772$  & $60964\pm4536$ \\
    1000 & 2.28 & $\mathbf{19917}\pm892$ & $48412\pm8630$& $186094\pm1944$ & $265476\pm4276$ & $272350\pm4978$ & $702645\pm54655$ \\
    \bottomrule
  \end{tabular}
\end{table}

The results are reported in Table~\ref{tab:eBAI_comparison}.
All methods achieve essentially perfect PGS on this benchmark, while ENDS consistently requires the smallest stopping budget across all values of $K$.
The advantage becomes more pronounced as the number of alternatives increases, suggesting that ENDS scales favorably in this unknown-variance setting.
We note, however, that ENDS is not designed for large-scale problems, where the computational burden can become substantial.

\subsection{Multi-fidelity R\&S}\label{sec:num_mf_bai}
We consider four synthetic instances.
In each instance, the configuration is specified by the fidelity-wise mean vectors
$\bm\theta_{\cdot,m}$, the bias vector $\bm\xi$, and the sampling cost vector $\bm C$.
The top fidelity determines the true best arm.
Instance~1 has $(K,M)=(5,2)$, $\bm\theta_{\cdot,1}=(0.4, 0.4, 0.4, 0.4, 0.5)$, and $\bm\theta_{\cdot,2}=(0.5, 0.5, 0.5, 0.5, 0.6)$, with $\bm\xi=(0.1, 0)$ and $\bm C=(0.5, 5)$.
In this instance, the optimal allocation balances sampling across the two fidelities.
Instance~2 has $(K,M)=(4,2)$, $\bm\theta_{\cdot,1}=(0.9, 0.4, 0.4, 0.3)$, and $\bm\theta_{\cdot,2}=(0.5, 0.6, 0.7, 0.85)$, with $\bm\xi=(0.55, 0)$ and $\bm C=(0.1, 5)$.
Here the low fidelity acts as a misleading trap, and the optimal allocation samples exclusively from the high fidelity.
Instance~3 has $(K,M)=(4,2)$, $\bm\theta_{\cdot,1}=(0.15, 0.25, 0.55, 0.85)$, and $\bm\theta_{\cdot,2}=(0.1, 0.2, 0.6, 0.8)$, with $\bm\xi=(0.1, 0)$ and $\bm C=(0.5, 5)$.
In this case, the optimal allocation samples exclusively from the low fidelity.
We also include Instance~4, randomly generated with $(K,M)=(5,4)$, following the setup in Table~2 of  \citet{poiani2024optimal}.
For all instances, we fix target confidence level $\delta = 0.05$ and Gaussian observation with known variance $\sigma^2=1$.

We compare ENDS with MF-GRAD \citep{poiani2024optimal}, TS-KKT-IDS \citep{qin2025dual} applied only at the highest fidelity, GRAD \citep{menard2019gradient} applied only at the highest fidelity, and two uniform allocation schemes: uniform sampling and uniform cost allocation.
To ensure a fair comparison, all algorithms use the same stopping rule in \eqref{eq:GLR-stopping} with a heuristic stopping threshold $\beta_{n,\delta} = \log\bigl((\log n + 1)/\delta\bigr)$.

\begin{table}[htbp]
\centering
\footnotesize
\caption{Comparison of expected stopping cost ($\pm$ Monte Carlo standard error) under confidence level $\delta=0.05$ and observation variance $\sigma^2=1$. PCS omitted, since all algorithms achieve a PCS of $>0.998$.}
\label{tab:mf_algo_comparison_0.1}\
\begin{tabular}{ccccccc}
\toprule
\textbf{Inst.} & \textbf{ENDS} & \textbf{MF-GRAD} & \textbf{TS-KKT-IDS} & \textbf{GRAD} & \textbf{Uniform Sample} & \textbf{Uniform Cost} \\
\midrule
1 & \textbf{32237} $\pm$ 531  & 65611 $\pm$ 1088 & 53841 $\pm$ 712  & 77602 $\pm$ 1086 & 56952 $\pm$ 814  & 67366 $\pm$ 1012 \\
2 & 11504 $\pm$ 199           & 38790 $\pm$ 671  & \textbf{10794} $\pm$ 203 & 17215 $\pm$ 352  & 18883 $\pm$ 347  & 38980 $\pm$ 656  \\
3 & \textbf{3550} $\pm$ 84    & 4231 $\pm$ 103   & 5085 $\pm$ 101   & 8938 $\pm$ 196   & 6844 $\pm$ 153   & 4539 $\pm$ 103   \\
4 & \textbf{3613} $\pm$ 65    & 3953 $\pm$ 78    & 21815 $\pm$ 330  & 28922 $\pm$ 455  & 11765 $\pm$ 184  & 3734 $\pm$ 69    \\
\bottomrule
\end{tabular}
\end{table}

We make several observations from these experiments.
First, ENDS achieves the lowest expected stopping cost in almost all instances while maintaining PCS equal to $1$.
The only exception is Instance~2, where it is slightly outperformed by TS-KKT-IDS.
In Instance~2, the optimal allocation places all mass on the highest fidelity, so restricting attention to that fidelity is advantageous; indeed, ENDS restricted to the highest fidelity coincides exactly with TS-KKT-IDS.
Second, although MF-GRAD is based on the same optimal allocation problem in \eqref{eq:Gamma_MF}, it treats sampling allocation as a continuous optimization problem followed by a tracking step.
By contrast, ENDS acts directly as a bandit sampling rule, which appears markedly more efficient in our experiments.
Third, Instance~2 shows that ENDS is robust to misleading low-fidelity information.
When the low fidelity distorts the arm ranking, MF-GRAD incurs much larger cost because it converges slowly to the optimal allocation that excludes the low fidelity.
ENDS instead performs nearly as well as TS-KKT-IDS, which effectively benefits from oracle knowledge that the low fidelity should be ignored.
Fourth, in Instance~4, most of the discriminative information lies in the lower fidelities.
This explains the poor performance of TS-KKT-IDS and GRAD, which use only the highest fidelity, and the competitive performance of uniform cost allocation, which assigns very little mass to the highest fidelity.
Finally, all algorithms achieve nearly perfect PCS, suggesting that the heuristic stopping rules are conservative.
Sharper threshold calibration is left for future work, see also \citet{ding2025stopping}.

\subsection{Dueling bandits}\label{sec:num_dueling}

We compare ENDS with state-of-the-art dueling bandit algorithms on random instances drawn from the family considered by \citet{haddenhorst2021identification}.
As baselines, we implement SAVAGE \citep{urvoy2013generic}, DKWT \citep{haddenhorst2021identification}, SEEBS \citep{ren2020sample}, Explore-then-Verify (EtV, \citealt{NIPS2016_65b9eea6}), and a uniform-allocation baseline.
For each algorithm, we averaged over $1000$ independently sampled instances, sampled uniformly from the set of $K$-armed preference matrices such that: (i) there is a unique Condorcet winner;
(ii) every off-diagonal entry is at least $h/2$ away from $1/2$.
The parameter $h$ controls the hardness of the instance, with smaller values corresponding to more difficult problems.
ENDS and Uniform-allocation use the GLR stopping \eqref{eq:GLR-stopping}, while all other algorithms use their own stopping rules.

\begin{table}[htbp]
  \centering
  \footnotesize
  \caption{Comparison of the expected stopping budget $\E[B_{\tau_\delta}]$ ($\pm$ Monte Carlo standard error) under different configurations of $(m,h)$ and $\delta = 0.05$. PCS omitted, since all algorithms achieve a perfect PCS of 1.}
  \label{tab:algo_comparison_db}
  \begin{tabular}{l*{6}{c}}
    \toprule
    \textbf{$(m,h)$} & \textbf{ENDS} & \textbf{SAVAGE} & \textbf{Uniform} & \textbf{DKWT} & \textbf{SEEBS} & \textbf{EtV} \\
    \midrule
    $(5,0.20)$  & $\mathbf{173} \pm 4$   & $953 \pm 15$    & $501 \pm 20$    & $6506 \pm 103$   & $7535 \pm 134$   & $8751 \pm 176$ \\
    $(5,0.15)$  & $\mathbf{206} \pm 5$   & $1246 \pm 24$   & $670 \pm 30$    & $8293 \pm 142$   & $11779 \pm 239$  & $11183 \pm 280$ \\
    $(5,0.10)$  & $\mathbf{303} \pm 10$  & $1825 \pm 47$   & $1058 \pm 57$   & $14198 \pm 396$  & $19404 \pm 472$  & $20464 \pm 1803^{\dagger 2}$ \\
    $(5,0.05)$  & $\mathbf{479} \pm 27$  & $3580 \pm 146$  & $2081 \pm 158$  & $31291 \pm 1255$ & $37198 \pm 1133$ & $50368 \pm 4220^{\dagger 14}$ \\
    $(10,0.20)$ & $\mathbf{295} \pm 4$   & $2805 \pm 22$   & $809 \pm 32$    & $14509 \pm 155$  & $17520 \pm 194$  & $26281 \pm 317$ \\
    $(10,0.15)$ & $\mathbf{336} \pm 5$   & $3276 \pm 33$   & $1065 \pm 57$   & $18475 \pm 200$  & $27447 \pm 347$  & $33373 \pm 730$ \\
    $(10,0.10)$ & $\mathbf{409} \pm 7$   & $4321 \pm 63$   & $1471 \pm 90$   & $32826 \pm 596$  & $45535 \pm 689$  & $58731 \pm 3330^{\dagger 7}$ \\
    $(10,0.05)$ & $\mathbf{494} \pm 11$  & $6888 \pm 177$  & $2701 \pm 446$  & $74341 \pm 1910$ & $81790 \pm 1567$ & $154396 \pm 8501^{\dagger 76}$ \\
    \bottomrule
    \multicolumn{7}{l}{\footnotesize $^{\dagger x}$ indicates that the algorithm was forcibly terminated at $10^6$ steps on $x$ out of the $1000$ instances.}
  \end{tabular}
\end{table}

Table~\ref{tab:algo_comparison_db} reports the expected stopping cost $\E[B_{\tau_\delta}]$ ($\pm$ Monte Carlo standard error).
Across all settings, ENDS is the most cost-efficient method.
The baselines with their native stopping rules incur substantially larger costs.
This comparison also suggests that the GLR-based stopping rule, which is decoupled from the sampling rule, can be considerably more efficient than elimination-type stopping rules designed jointly with their sampling rules.
Thus, the observed advantage should not be attributed solely to the ENDS sampling rule; it is partly intertwined with the use of a stronger stopping rule.
The same caveat applies to the comparisons with KN, Paulson, and ZH-Proc2 (but not KE-Alg2) in Section~\ref{sec:num_good_alternative}.
In contrast, all algorithms in Section~\ref{sec:num_mf_bai} use the same stopping rule.

\section{Conclusion}

We studied cost-aware fixed-precision ranking-and-selection problems in which the correct answer may be non-unique and intermediate estimates may be non-answerable.
To handle this setting, we proposed a unified framework based on answer-wise acceptance sets, restricted generalized likelihood ratio stopping, and an answer--pitfall decomposition, which leads to a common max-max-min characteristic value and the ENDS procedure.
We derived explicit instantiations for good-alternative selection, multi-fidelity ranking-and-selection, and Condorcet-winner dueling bandits, and numerical experiments suggest that this unified recipe is robust and competitive across these qualitatively different problems.

Several directions remain open.
A natural next step is to establish stronger theoretical guarantees for ENDS, such as asymptotic, instance-dependent upper bounds that match the characteristic value.
A second is to develop a sharper understanding of the stopping rule, including less conservative threshold calibration and nonasymptotic $\delta$-correctness guarantees.
It would also be interesting to extend the framework to broader structured settings with richer feedback models, such as contextual problems.

\section*{Acknowledgments}
Wei You's research is generously supported by the Hong Kong Research Grants Council [Grant GRF 16212823] and [Theme-based Research Project T32-615/24-R].

\bibliographystyle{plainnat}
\bibliography{refs}

\end{document}